\documentclass[10pt,twocolumn,letterpaper]{article}

\usepackage{iccv}
\usepackage{times}
\usepackage{epsfig}
\usepackage{graphicx}
\usepackage{amsmath}
\usepackage{amssymb}
\usepackage{caption}
\usepackage{array} 
\usepackage{multirow}
\usepackage[ruled,linesnumbered]{algorithm2e}
\usepackage{authblk}

\graphicspath{{figures/}}

\usepackage[pagebackref=true,breaklinks=true,letterpaper=true,colorlinks,bookmarks=false]{hyperref}

\iccvfinalcopy 

\def\iccvPaperID{10994} 

\ificcvfinal\fi

\begin{document}

\title{Meta-Auxiliary Network for 3D GAN Inversion}

\author[1]{Bangrui Jiang}
\author[2]{Zhenhua Guo}
\author[1]{Yujiu Yang}
\affil[1]{Tsinghua Shenzhen International Graduate School, Tsinghua University, China}
\affil[2]{Tianyi Traffic Technology, China}


\maketitle
\ificcvfinal\thispagestyle{empty}\fi

\begin{abstract}
Real-world image manipulation has achieved fantastic progress in recent years. GAN inversion, which aims to map the real image to the latent code faithfully, is the first step in this pipeline. However, existing GAN inversion methods fail to achieve high reconstruction quality and fast inference at the same time. In addition, existing methods are built on 2D GANs and lack explicitly mechanisms to enforce multi-view consistency.
In this work, we present a novel meta-auxiliary framework, while leveraging the newly developed 3D GANs as generator. The proposed method adopts a two-stage strategy. In the first stage, we invert the input image to an editable latent code using off-the-shelf inversion techniques. The auxiliary network is proposed to refine the generator parameters with the given image as input, which both predicts offsets for weights of convolutional layers and sampling positions of volume rendering. In the second stage, we perform meta-learning to fast adapt the auxiliary network to the input image, then the final reconstructed image is synthesized via the meta-learned auxiliary network. Extensive experiments show that our method achieves better performances on both inversion and editing tasks.
\end{abstract}

\section{Introduction}

\label{sec:intro}

In recent years, Generative Adversarial Networks (GANs) \cite{goodfellow2020generative} have been significantly developed and achieve remarkable performance, which can generate images with both high-resolution and high-quality \cite{karras2017progressive,karras2019style}. Moreover, numerous studies \cite{alaluf2021only, collins2020editing, patashnik2021styleclip} have illustrated that the latent space learned by these models encodes a broad range of meaningful semantics, enabling the manipulation of synthesized images. Consequently, comprehending and investigating a well-trained GAN model constitutes a crucial and active research domain.

To edit real-world images, a widely adopted approach involves a two-stage process comprising GAN inversion and latent code editing. GAN inversion is the initial step, where an image is inverted into the latent space of a GAN model to find the corresponding latent code. Following this, the latent code is edited in a semantically meaningful manner to obtain a new code that is used to generate the edited output image. The existing works \cite{richardson2021encoding,alaluf2022hyperstyle,dinh2022hyperinverter} primarily focus on GAN inversion, aiming to reconstruct an image from the latent code that closely resembles the input image. 
When performing GAN inversion, the objective is not just to faithfully reconstruct the input image, but also to enable effective image editing in the subsequent stages. However, previous studies \cite{tov2021designing} have highlighted the existence of a trade-off between image reconstruction and editing. This trade-off is known to depend primarily on the embedding space where the input image is mapped to. For instance, StyleGAN \cite{karras2019style,karras2020analyzing} contains two popular embedding spaces, namely the native StyleGAN $\mathcal{W}$ space and the extended $\mathcal{W}^+$ space. In general, inverting an image to the $\mathcal{W}$ space yields excellent editability. However, it has been shown to be infeasible for faithfully reconstructing the input image. Conversely, the $\mathcal{W}^+$ space enables more accurate reconstructions, but it is associated with limited editing capabilities.

As for GAN inversion, the most recent works \cite{roich2022pivotal,dinh2022hyperinverter,alaluf2022hyperstyle} adopt two-stage strategy to achieve high reconstruction accuracy. These methods first use off-the-shelf methods \cite{abdal2019image2stylegan,richardson2021encoding} to determine an approximate latent code. Then, they augment the latent space to include the given image by slightly altering the generator. PTI \cite{roich2022pivotal} directly fine-tune the generator via hundreds of optimization steps, which is time-consuming. HyperInverter \cite{dinh2022hyperinverter} leverages the hypernetworks to predict the residual weights of the generator, which reduces processing time but sacrifices reconstruction accuracy. Moreover, existing methods often results in unpredictable changes to appearance or identity when editing pose, due to the lack of modeling the 3D structure.

In this work, we address these limitations to achieve multi-view-consistent image editing while obtaining robust reconstruction accuracy and high inference speed. Our method leverages recently developed 3D-aware GAN, i.e. EG3D \cite{chan2022efficient}, as generator. Since EG3D is designed with the StyleGAN backbone from the ground up, it inherits the well-studied properties of the StyleGAN latent space. Similar to previous work \cite{roich2022pivotal,dinh2022hyperinverter,alaluf2022hyperstyle}, our method also adopts a two-stage strategy. In the first stage, we invert the input image to an editable latent code using off-the-shelf inversion techniques. In addition, the auxiliary network is proposed to refine the generator parameters with the given image as input. The auxiliary network consists of two part: one predicts offsets for the weights of the convolutional layers which can recover the lost details, the other predicts offsets for sampling positions of volume rendering which can rectify structural errors. In the second stage, we perform meta-learning to adapt the auxiliary network to the input image, then the final reconstructed image is synthesized via the updated auxiliary network. Different from PTI which directly fine-tune the generator, our meta-auxiliary network can adapt to a new image in few steps which significantly reduces processing time and maintains comparable performance.

The main contributions can be summarized as follows:
\begin{itemize}
\item We present a 3D-aware GAN framework for GAN inversion and image editing. A novel auxiliary network is proposed to update the parameters of a pre-trained generator and rectify volume rendering process.

\item We are the first to incorporate meta-learning into GAN inversion. With meta-learning strategy, the auxiliary network can quickly adapt to unseen images.

\item Experimental results demonstrate the superior performance of our method compared to the existing works.

\end{itemize}

\section{Related  Work}
\noindent\textbf{Latent Space Manipulation}\quad 
The latent space of a well-trained GAN generator encapsulates numerous interpretable semantic directions, which can be utilized for image editing. Hence, multiple methods have been proposed for discovering semantic latent directions through different levels of supervision. Some methods \cite{abdal2021styleflow, goetschalckx2019ganalyze, shen2020interpreting} leverage semantic labels for full supervision, necessitating pre-trained attribute classifiers and being limited to known attributes. Other methods \cite{harkonen2020ganspace, shen2021closed, voynov2020unsupervised, yuksel2021latentclr} adopt principal component analysis or contrastive learning to explore unique editing directions in an unsupervised manner. However, in order to apply latent manipulation to real images, GAN inversion should be first performed.

\begin{figure*}[t]
\centering
\begin{minipage}{1.0\textwidth}
\centering
\includegraphics[width=1.0\textwidth]{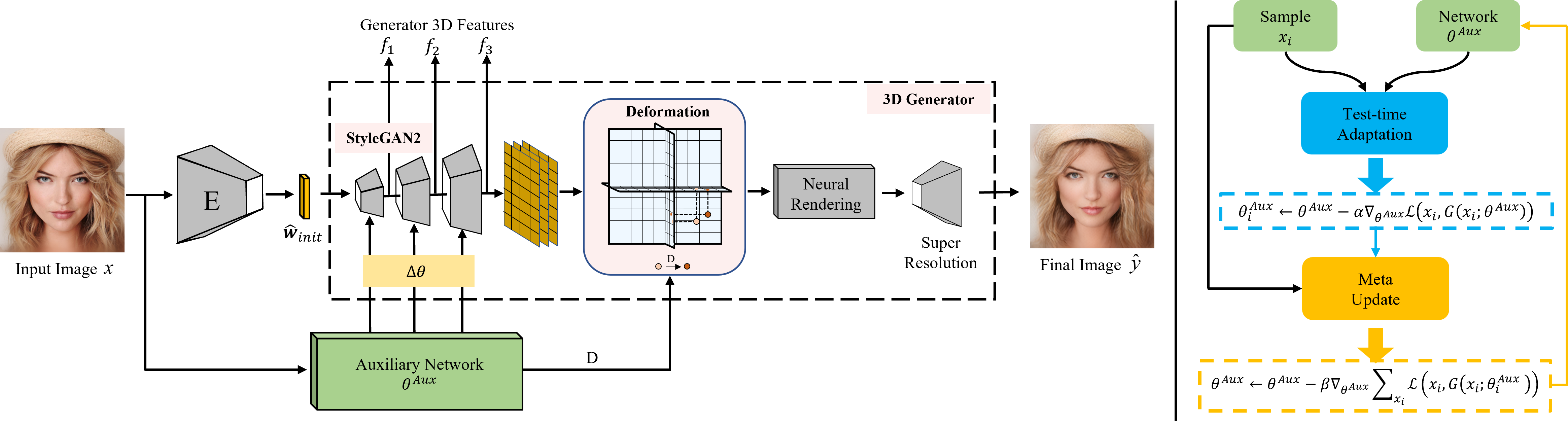}
\end{minipage}
\caption{Given an image $x$, we begin with an approximate latent code $\hat{\omega}_{init}$ predicted by an off-the-shelf encoder. With the image $x$ as input, our auxiliary network predicted a set of offsets ${\Delta}{\theta}$, which are used to modulate both StyleGAN's parameters and volume rendering. To assist the auxiliary network, we propose a meta-learning training scheme for fast adaptation, which can update auxiliary network during inference. With the meta-learned auxiliary network, the generator $G$ will be parameterized with new params $\hat{\theta}^G$ and generated final output $\hat{y}$.}
\label{fig:whole}
\end{figure*}

\noindent\textbf{GAN Inversion}\quad 
GAN inversion \cite{zhu2016generative} is the process of locating a latent code that can be passed to the generator to reconstruct a given image. Existing methods can be roughly categorized into three groups: optimization-based, encoder-based and two-stage methods. Optimization-based methods \cite{abdal2019image2stylegan,abdal2020image2stylegan++,creswell2018inverting,kang2021gan} directly optimize the latent code to minimize the reconstruction error for a given image, which are time-consuming but achieve high accuracy. Encoder-based methods \cite{alaluf2021restyle,pidhorskyi2020adversarial,richardson2021encoding,wang2022high,hu2022style} train an encoder over a large number of samples to learn a mapping from an image to its latent representation. Such methods are efficient during inference but inferior in reconstruction quality to optimization-based method. Some two-stage methods \cite{zhu2020domain,bau2019seeing} combine both above approaches, which first encode images to initial latent codes and then optimize the latent codes. Instead of optimizing the latent codes, other two-stage methods \cite{roich2022pivotal,alaluf2022hyperstyle,dinh2022hyperinverter} turn to fine-tune the generator. HyperStyle \cite{alaluf2022hyperstyle} and HyperInverter \cite{dinh2022hyperinverter} utilize an additional hypernetwork to refine the generator weights with respect to the given image. PTI \cite{roich2022pivotal} directly adopt backpropagation for weight optimization, which achieves the best reconstruction performance but requires a substantially longer time. In comparison to PTI, our method adopts meta learning to accelerate optimization process and improve reconstruction quality.

\noindent\textbf{Meta-Learning}\quad 
To achieve test-time adaptation without greatly increasing the cost of computation, meta-learning has been proved to be effective. 
Meta-learning is originally proposed in few-shot classification, which aims to learn prior knowledge across tasks \cite{lee2019meta,baik2021meta,goldblum2020unraveling}. Among the meta-learning systems, MAML \cite{finn2017model} has greatly enjoyed the attention for its simplicity and generalizability.
Recently, several works \cite{park2020fast,soh2020meta,chi2021test,choi2020scene} extent the MAML scheme to low-level vision tasks. It allows the pre-trained model to be optimized in a way such that it can quickly adapt to any given image. The generalizability of its model-agnostic algorithm motivates us to integrate test-time adaptation into GAN inversion using MAML training scheme.

\noindent\textbf{Generative 3D-aware image synthesis}\quad 
Building on the success of 2D image-based GANs \cite{karras2019style,karras2020analyzing}, recent efforts have focused on training 3D-aware multi-views consistent GANs from collections of single-view 2D images in an unsupervised manner. Achieving this challenging goal requires a combination of a neural scene representation and differentiable rendering algorithm. Recent work in this domain builds on representation using meshes \cite{liao2020towards}, voxel grids \cite{hao2021gancraft}, multiple planes \cite{deng2022gram,chan2022efficient}, or a combination of low-resolution voxel grids with 2D CNN-based layers \cite{niemeyer2021giraffe,gu2021stylenerf}. 
Among these methods, the current SOTA method EG3D \cite{chan2022efficient} uses an tri-plane-based volume representation combined with volume rendering. 
In this study, we extend GAN inversion from 2D-based StyleGAN to 3D-based EG3D, which allows the generator to synthesize images in multiple views and apply latent space manipulation.

\section{Method}

\subsection{Preliminaries}

GAN inversion task aims to optimize a latent
code ${\omega}$ that can be passed to the pre-trained generator to reconstruct a given image $x$:

\begin{equation}
\label{equ:encoder}
\hat{\omega} = \mathop{\arg\min}\limits_{\omega} \mathcal{L}(x, G(\omega;\theta^G))
\end{equation}
where $G(\omega;\theta^G)$ is the image reconstructed by a pre-trained generator $G$ with parameter $\theta^G$, using the latent code $\omega$. $\mathcal{L}$ is the loss objective. Solving Eq.\ref{equ:target} via optimization-based methods \cite{abdal2019image2stylegan,abdal2020image2stylegan++,creswell2018inverting,kang2021gan} typically requires hundreds of iterations, which takes several minutes per image. To improve the performance, encoder-based methods introduce an encoder $E$ to predict latent code as $\hat{\omega}=E(x)$, which only takes a few seconds for inference. Then, a latent manipulation $f$ can be applied over the inverted latent code $\hat{\omega}$ to obtain an edited image as $G(f(\hat{\omega});\theta^G)$. In practice, the commonly used latent code manipulation method is InterFaceGAN \cite{shen2020interpreting}, and the manipulation $f$ can be formulated as $f(\hat{\omega}) = \hat{\omega} + \alpha n$, where $\alpha$ is magnitude constant and $n$ represents semantic direction of a specific facial attribute.

Apart from finding more accurate latent code $\hat{\omega}$, recent works \cite{roich2022pivotal,dinh2022hyperinverter,alaluf2022hyperstyle} turn to inject new identities into the well-behaved latent space of generator. Given a target image, they first utilize existing methods to find an initial latent code $\hat{\omega}_{init}$ leading to an approximate reconstruction. Then, either an optimization process or a hypernetwork is adopted to adapt the generator parameters to the specific image:

\begin{equation}
\label{equ:target}
\hat{\theta}^G = \mathop{\arg\min}\limits_{\theta^G} \mathcal{L}(x, G(\hat{\omega}_{init};\theta^G))
\end{equation}
where $\hat{\theta}^G$ represents the adapted generator parameters. The final reconstruction image is obtained by utilizing the initial latent and adapted parameters as $\hat{y}=G(\hat{\omega}_{init};\hat{\theta}^G)$.

\begin{figure}[tbp]
\centering

\begin{minipage}{1.0\linewidth}
\centering
\includegraphics[width=1.0\linewidth]{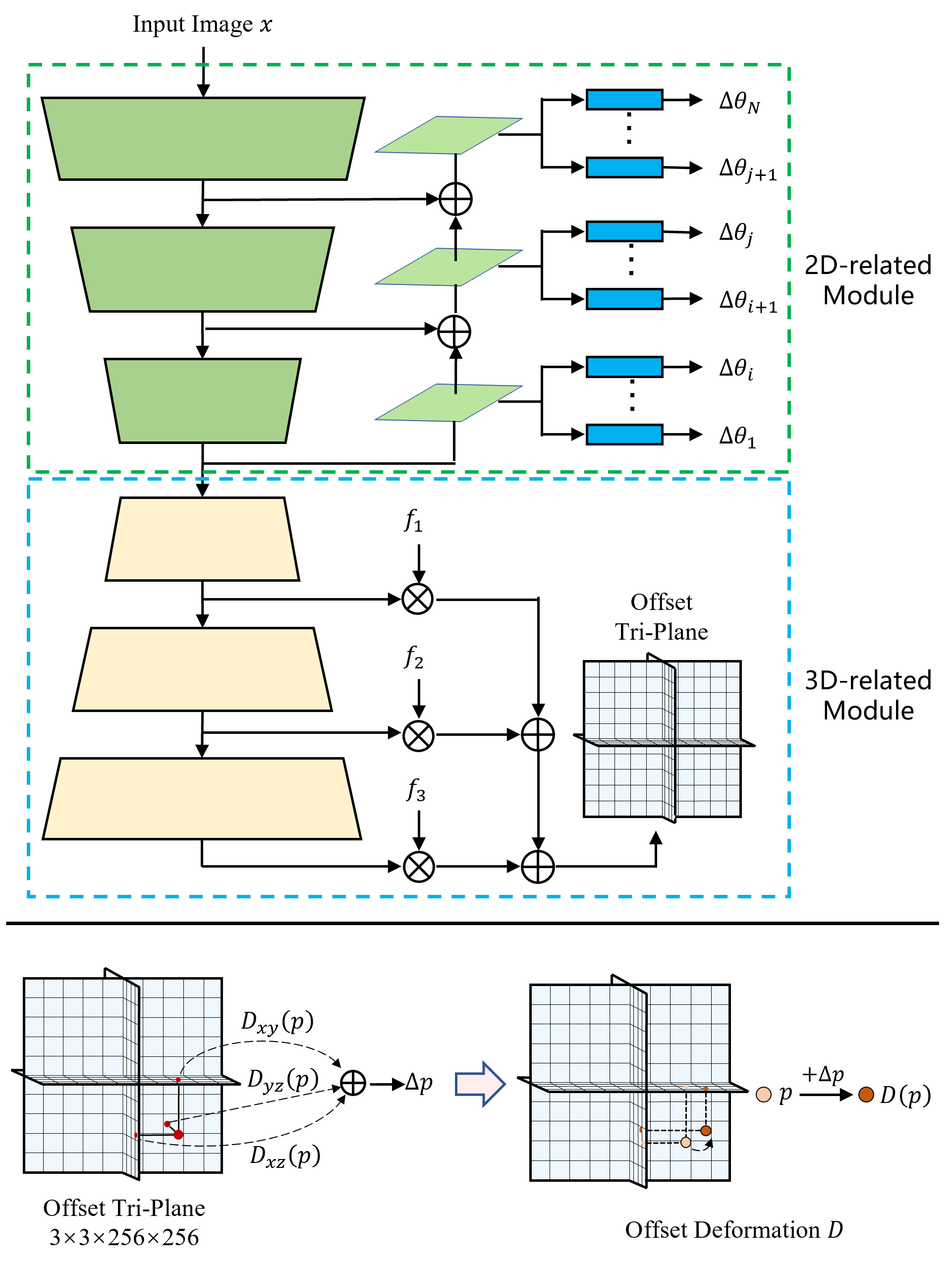}
\end{minipage}

\caption{Auxiliary network design. The network consists of 2D-related module and 3D-related module. The network first encode the input image into features. Then the 2D-related module predicts the residual weights $\Delta \theta$ for convolutional layers, while the 3D-related module predicts an offset tri-plane structure for coordinate deformation in volume rendering.}
\label{fig:architecture}
\end{figure}

\subsection{Overview}
Our proposed method is designed to efficiently modify the parameters of the generator in order to adapt to unseen images, as illustrated in Fig. \ref{fig:whole}. The input of our method includes an image $x$, a generator $G$ with parameters $\theta^G$, and an initial inverted latent code $\hat{\omega}_{init}$, which is obtained using an off-the-shelf encoder \cite{hu2022style}.

To achieve our goal of minimizing the objective defined in Eq. \ref{equ:target}, we introduce an auxiliary network $Aux$ that is responsible for predicting a new set of parameters $\hat{\theta}^G$ for generator $G$. The predicted parameters are given by $\hat{\theta}^G=Aux(x;\theta^{Aux})$, where $\theta^{Aux}$ are parameters of the auxiliary network. However, the auxiliary network only takes the target image $x$ as input without the initial output $G(\hat{\omega}_{init};\theta^G)$, which is insufficient to infer the desired modifications. Therefore, to assist the auxiliary network, we introduce meta-learning strategy to fine-tune the auxiliary network for test-time adaptation. 

To balance the trade-off between reconstruction and editability, it is crucial that the initial latent code resides in a well-behaved region of the latent space. To achieve this, we utilize the encoder to invert the input image into the $\mathcal{W}$ space, and the encoder is kept fixed during the training process. 
It will be demonstrated that although making adjustments around the initial latent code, the same editing methods as those used with the original generator can be applied.

In practice, rather than directly predicting the generator parameters, our auxiliary network predicts a set of offsets with respect to the original parameters. In addition, to balance between fidelity and efficiency, we perform meta-learning with a small number of iterations (5 in this work).

\subsection{Model Architecture}
The goal of auxiliary network is to provide additional information which can not be recovered by the latent code. The detailed architecture of the auxiliary network is shown in Fig. \ref{fig:architecture}, which can be divided into two modules.

The 2D-related module is to provide the missing texture information. We first employ a ResNet-101 \cite{he2016deep} to extract intermediate features from the input image. Inspired by HyperStyle \cite{alaluf2022hyperstyle}, a set of hypernetworks is used to predict a set of residual weights. We then use the residual weights to update the corresponding convolutional layers in the generator. Assume that the pre-trained generator has $N$ convolutional layers with weights $\theta_{conv}=(\theta_1, \dots, \theta_N)$. We therefore propose to use several small hypernetworks $H_j$ to predict the residual weights $\Delta\theta_j$ of each convolutional layer, where $j \in \{1,\dots,N\}$. Finally, the updated generator has the convolutional weights as $\hat{\theta}_{conv}=(\theta_1+\Delta\theta_1, \dots, \theta_N+\Delta\theta_N)$.

The 3D-related module aims to rectify unaligned structure. The pre-trained generator leverage the tri-plane feature representation, which consists three axis-aligned 2D feature planes. The feature of any 3D point $\mathbf{p} \in \mathbb{R}^3$ is queried by projecting $\mathbf{x}$ onto the planes, retrieving three feature vectors via bilinear interpolation, and aggregating the vectors by summation, i.e., $F(\mathbf{p})=F_{xy}(\mathbf{p})+F_{yz}(\mathbf{p})+F_{xz}(\mathbf{p})$ where $F_{ij}:\mathbb{R}^3\mapsto\mathbb{R}^C$ is a function mapping 3D coordinates to features on the $ij$ plane via projection and interpolation. The subsequent volume rendering will infer the corresponding feature $F(\mathbf{p})$ with sampling point $\mathbf{p}$.

Inspired by \cite{bergman2022generative}, we introduce a deformation function $D: \mathbb{R}^3 \mapsto \mathbb{R}^3$ to rectify coordinate $\mathbf{p}$. Then, the corresponding feature of $\mathbf{p}$ is formulated as
\begin{equation}
F(\mathbf{p})=(F_{xy}\circ D)(\mathbf{p})+(F_{yz}\circ D)(\mathbf{p})+(F_{xz}\circ D)(\mathbf{p})
\end{equation}

In practice, we construct a similar offset tri-plane structure to realize the deformation function $D$. As shown in Fig. \ref{fig:architecture}, we take the highest feature from 2D-related module as input followed by several StyleGAN blocks, which is modulated by the viewpoint of the input image. In this sense, the 3D-related module can reconstruct 3D features from input image. Therefore, the structural offsets can be inferred by comparing 3D features from input image and generator, which constitutes an offset tri-plane structure. Then, the deformation $D$ can be written as
\begin{equation}
\Delta \mathbf{p}=D_{xy}(\mathbf{p})+D_{yz}(\mathbf{p})+D_{xz}(\mathbf{p})
\end{equation}
where $D_{ij}:\mathbb{R}^3\mapsto\mathbb{R}^3$ is a function mapping 3D coordinates to offsets and the deformed coordinate can be written as $D(\mathbf{p})=\mathbf{p}+\Delta \mathbf{p}$. Consequently, the adapted generator parameters $\hat{\theta}^G$ consist of both adapted weights $\hat{\theta}_{conv}$ and sampling point deformation $D$.

\SetKwInOut{KwIn}{Require}
\begin{algorithm}[!t]
  \SetAlgoLined
  \KwIn{$p(x)$: uniform distributions over images}
  \KwIn{$\alpha,\beta$: step size hyper-parameters}

  Initialize parameters $\theta^{Aux}$\;
  \While{not converged}{
    Sample a batch of images $x_i \sim p(x)$\;
    \ForEach{i}{
    Evaluate $\nabla_{\theta^{Aux}} \mathcal{L}(x_i, G(x_i; \theta^{Aux}))$\;
    Compute adapted parameters using \;
    $\theta^{Aux}_i \leftarrow \theta^{Aux} - \alpha \nabla_{\theta^{Aux}} \mathcal{L}(x_i, G(x_i; \theta^{Aux}))$
    }
    $\theta^{Aux} \leftarrow \theta^{Aux} - \beta \nabla_{\theta^{Aux}} \sum\limits_{x_i} \mathcal{L}(x_i, G(x_i; \theta^{Aux}_i))$
    }
  \caption{Meta learning algorithm}
  \label{al1}
\end{algorithm}

\subsection{Meta-Learning Strategy}
The results obtained from the above architecture is sub-optimal since it only exploits the external data and does not take advantage of internal information from test images. Therefore, we introduce meta-learning strategy to learn model parameters to facilitate test-time adaptation. For a test image, our auxiliary network is update and adapted to the specific test image. It is necessary to be noted that since the pre-trained generator contains diverse facial information, directly applying meta-learning strategy to the generator's parameters would damage the prior information.

In this paper, we adopt the model-agnostic meta-learning (MAML) \cite{finn2017model} approach. MAML can find a good initialization of the parameters that are sensitive to changes in task, so that small update can make large improvements. The entire training procedure is listed in Alg. \ref{al1}. For one gradient update, new adapted parameters is

\begin{equation}
\label{equ:task}
\theta^{Aux}_i = \theta^{Aux} - \alpha \nabla_{\theta^{Aux}} \mathcal{L}(x, G(x; \theta^{Aux}))
\end{equation}
where $\alpha$ is the task-level learning rate. In training stage, the model parameters $\theta^{Aux}$ are optimized to achieve minimal test error. Concretely, the meta-objective is
\begin{equation}
\label{equ:meta}
\mathop{\arg\min}\limits_{\theta^{Aux}} \sum_{x_i \sim p(x)} \mathcal{L}(x_i, G(x_i; \theta^{Aux}_i))
\end{equation}

Meta-learning optimization is performed using Eq. \ref{equ:meta}, which is to learn the knowledge across task. Any gradient-based optimization can be used for meta-learning training. 
For stochastic gradient descents, the parameter update rule is expressed as 
\begin{equation}
\label{equ:update}
\theta^{Aux} = \theta^{Aux} - \beta \nabla_{\theta^{Aux}} \sum_{x_i \sim p(x)} \mathcal{L}(x_i, G(x_i; \theta^{Aux}_i))
\end{equation}
where $\beta$ is the meta-learning rate. The above process is the training phase of meta-learning, which optimizes the auxiliary network so that it learns how to adapt to unseen samples. During the inference phase, only task-level update shown in Eq. \ref{equ:task} is performed.

\subsection{Training Loss}
Similar to previous methods, our training is guided by an image-space reconstruction objective. Particularly, the final loss objective is defined as:
\begin{equation}
\mathcal{L} = \mathcal{L}_2+\lambda_{lpips}\mathcal{L}_{LPIPS}+\lambda_{id}\mathcal{L}_{id}+\lambda_{adv}\mathcal{L}_{adv}
\label{equ:loss}
\end{equation}
where $\lambda_{lpips}$, $\lambda_{id}$, $\lambda_{adv}$ are constants. The pixel-wise loss $\mathcal{L}_2$ is defined as:
\begin{equation}
\mathcal{L}_2(x, \hat{y}) = \| x - \hat{y} \|_2
\end{equation}

The LPIPS \cite{zhang2018unreasonable} loss utilize the perceptual feature extractor $P$ to learn perceptual similarities, which is defined as:
\begin{equation}
\mathcal{L}_{LPIPS}(x, \hat{y}) = \| P(x) - P(\hat{y}) \|_2
\end{equation}

The identity loss uses the pre-trained ArcFace \cite{deng2019arcface} network $R$ to measure cosine similarity:
\begin{equation}
\mathcal{L}_{id}(x, \hat{y}) = 1 - \langle R(x) - R(\hat{y}) \rangle
\end{equation}
where $\langle \cdot \rangle$ denotes cosine similarity. And the adversarial loss $\mathcal{L}_{adv}$ is constructed following StyleGAN.

\section{Experiment}

\begin{table}[]
\centering
\setlength{\tabcolsep}{1.2mm}{
\begin{tabular}{l|ccc|c}
\hline
Method           & MSE$\downarrow$& LPIPS$\downarrow $& FID$\downarrow $& Time(s)$\downarrow $\\ \hline
pSp\cite{richardson2021encoding} & 0.0250    &  0.1500     & 25.00     &    1.72      \\
StyleTransformer\cite{hu2022style} & 0.0249    & 0.1501       & 26.81    & 1.56         \\
HyperInverter\cite{dinh2022hyperinverter} &   0.0206  & 0.1313      & 18.92    & 2.22          \\
PTI\cite{roich2022pivotal} & 0.0135    &    0.1011   & 18.48    & 96.23         \\
Ours             & 0.0148    & 0.0718      & 14.57    & 4.11         \\ \hline
\end{tabular}
}
\caption{Quantitative results for image reconstruction.}
\label{tab:quantitative}
\vspace{-10pt}
\end{table}

\begin{figure*}[!t]
\centering
\begin{minipage}{1.0\textwidth}
\centering
\includegraphics[width=1.02\textwidth]{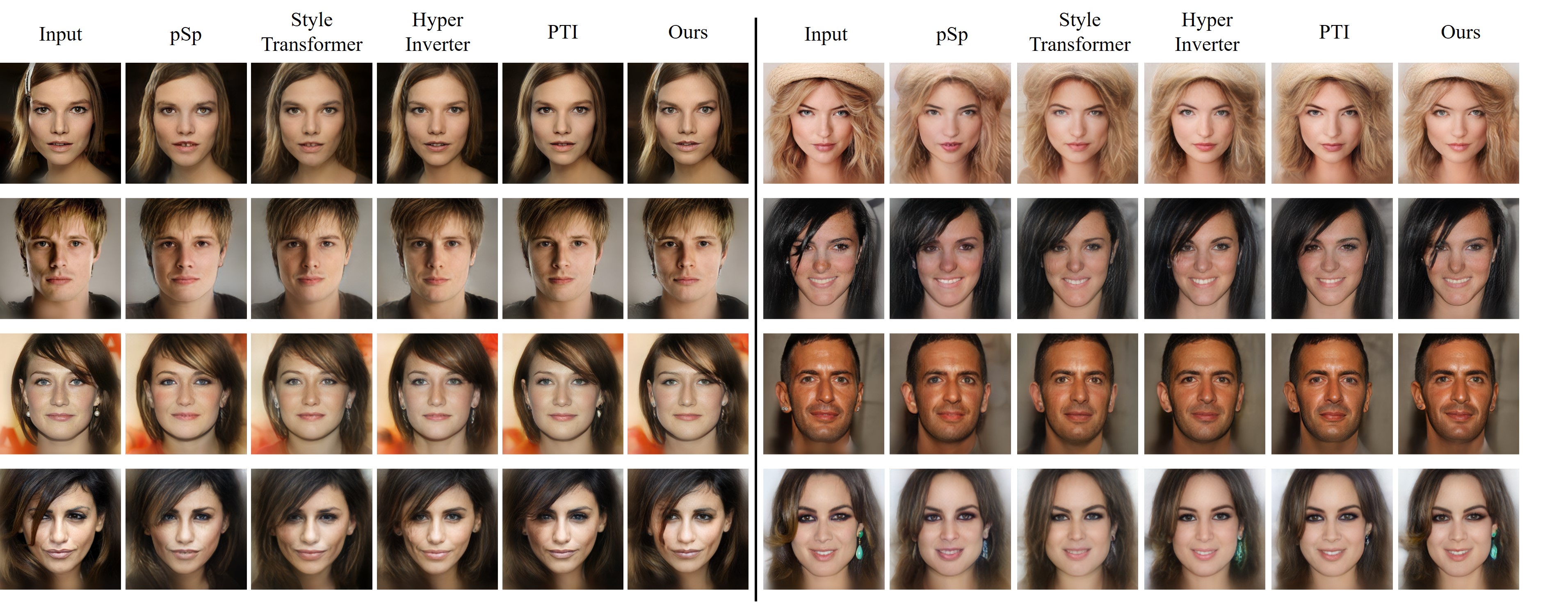}
\end{minipage}
\caption{Qualitative reconstruction comparison results.}
\label{fig:qualitative}
\end{figure*}

\subsection{Experimental Settings}
\noindent\textbf{Datasets and Baselines}\quad 
We conduct the experiments on the FFHQ \cite{karras2019style} as training dataset, and the CelebA-HQ \cite{karras2017progressive} as testing dataset. We compare our results with the state-of-the-art encoder-based methods pSp \cite{richardson2021encoding} and StyleTransformer \cite{hu2022style}. For two-stage methods, we choose PTI \cite{roich2022pivotal} and HyperInverter \cite{dinh2022hyperinverter} to compare with our results. It is worth noted that pSp and StyleTransformer employ $\mathcal{W}^+$ space, while PTI and HyperInverter choose $\mathcal{W}$ space. Specially, the above methods are originally implemented on StyleGAN2 \cite{karras2019style}. As EG3D \cite{chan2022efficient} is designed with StyleGAN2 backbone from the ground up, the above methods can be directly employed with EG3D as generator. Therefore, for fair comparisons, we use the official configurations to fine-tune these methods on EG3D.

\noindent\textbf{Implementation Details}\quad
In our experiments, the pre-trained EG3D generators being used are obtained directly from EG3D \cite{chan2022efficient} repository. To stabilize the meta-learning scheme, we divide the whole training procedure into two phase. In the first phase, we train the auxiliary network directly and obtain approximate parameters. In the second phase, we use the above parameter as initialization and train with meta-learning. For constants in the loss objective, we set $\lambda_{lpips}=0.8$, $\lambda_{id}=0.1$ and $\lambda_{adv}=0.005$.

\subsection{Reconstruction Results}
\noindent \textbf{Quantitative Results} \quad
We use several metrics to measure the reconstruction quality of our method compared with existing, including pixel-wise MSE, perceptual LPIPS \cite{zhang2018unreasonable} and FID \cite{heusel2017gans} as shown in Tab. \ref{tab:quantitative}. Compared with encoder-based methods (pSp and StyleTransformer), our method is able to reconstruct out-of-domain visual details, such as clothing and hairstyles. Although two-stage method PTI also utilizes optimization technique and achieve accurate reconstructions, PTI requires hundreds of optimization steps and comes with a high computational cost. On the contrary, our method can obtain comparable performance in few iteration steps, which demonstrates the effectiveness of meta-learning. In addition, the encoder-based two-phase method HyperInverter can not fully utilize the information to update the generator, while our method significantly outperform HyperInverter using optimization. Our method successfully reduce efficiency gap for encoder-based and optimization-based methods.

\noindent \textbf{Qualitative Results} \quad
We visualize the reconstruction results in Fig. \ref{fig:qualitative}. Compared with optimization-based method PTI, our method can achieve visually comparable results with an inference time several orders of magnitude faster. Further, compared with encoder-based method, our method better preserve texture details and better capture the input identity, such as hairstyles and clothing.

\begin{figure*}[!t]
\centering
\begin{minipage}{1.0\textwidth}
\centering
\includegraphics[width=1.0\textwidth]{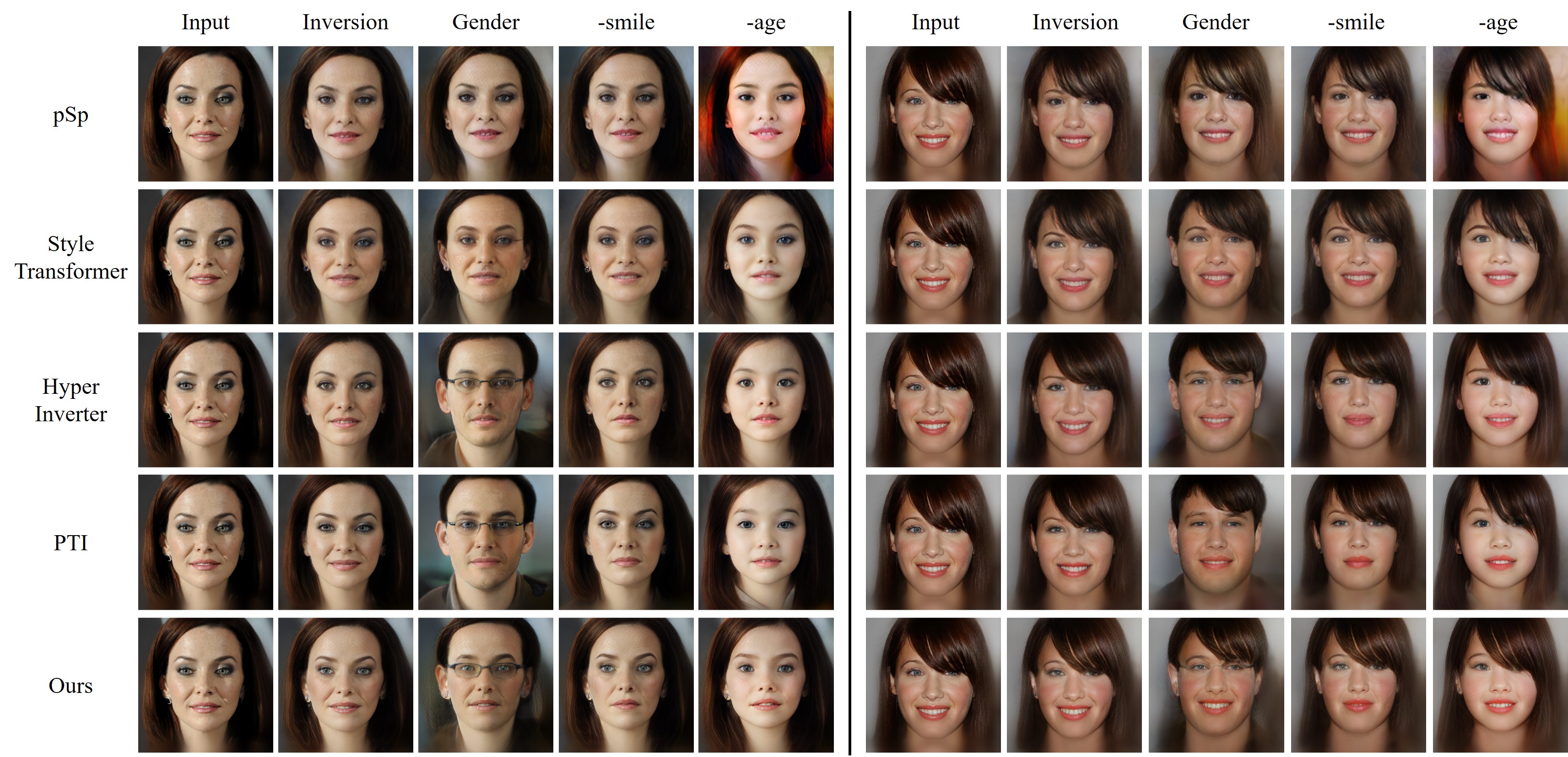}
\end{minipage}
\caption{Qualitative editing comparison results.}
\label{fig:edit}
\end{figure*}

\subsection{Editing Results}
\noindent \textbf{Quantitative Results} \quad
We first perform a quantitative evaluation for editing ability. Following previous works \cite{roich2022pivotal,dinh2022hyperinverter}, we present an experiment to test the effect of editing operator with the same editing magnitude on the latent code inverted by different inversion methods. We opt for age and gender for two editing directions in this experiment. Given the latent code $\omega$, we apply the editing operator to obtain the new latent code as $\omega_{edit}=\omega+\alpha*n$ where $\alpha$ is the editing magnitude and $n$ is the semantic direction learned by InterFaceGAN \cite{shen2020interpreting}. To quantitatively evaluate the editing ability, we measure the amount of age change for age edit and gender edit when applying the same $\alpha$ on each baseline. The result are shown in Tab. \ref{tab:edit}. Since our method works on highly editable $\mathcal{W}$ space, the editability of our method outperform single-stage encoder-based method. In addition, our method achieve comparable performance compared with other $\mathcal{W}$ space methods.

\begin{table}[]
\centering
\setlength{\tabcolsep}{1.1mm}{
\begin{tabular}{c|c|ccccc}
\hline
Dir. &
  \multicolumn{1}{c|}{} &
  \multicolumn{1}{c}{pSp} &
  \multicolumn{1}{c}{\begin{tabular}[c]{@{}c@{}}Style\\ Transformer\end{tabular}} &
  \multicolumn{1}{c}{\begin{tabular}[c]{@{}c@{}}Hyper\\ Inverter\end{tabular}} &
  \multicolumn{1}{c}{PTI} &
  Ours \\ \hline
\multirow{6}{*}{Age} &
  \multicolumn{1}{c|}{-3} & \multicolumn{1}{c}{-1.004} &-4.036 &\multicolumn{1}{c}{-5.247} &  \multicolumn{1}{c}{-4.619} &-4.365 \\
                                             & \multicolumn{1}{c|}{-2} & \multicolumn{1}{c}{-0.716} &-2.819  & \multicolumn{1}{c}{-3.646} & \multicolumn{1}{c}{-3.197} &-2.974  \\
                                             & \multicolumn{1}{c|}{-1} & \multicolumn{1}{c}{-0.393} &-1.492  & \multicolumn{1}{c}{-1.921} & \multicolumn{1}{c}{-1.985} & -1.544 \\
                                             & \multicolumn{1}{c|}{1}  & \multicolumn{1}{c}{0.390} &1.598  & \multicolumn{1}{c}{2.083} & \multicolumn{1}{c}{1.711} &1.645  \\
                                             & \multicolumn{1}{c|}{2}  & \multicolumn{1}{c}{0.868} &3.354  & \multicolumn{1}{c}{4.514} & \multicolumn{1}{c}{4.135} &3.701  \\
                                             & 3                       & 1.356                     &5.491  & 7.273                     &     7.342                 &5.716  \\ \hline
\multicolumn{1}{l|}{\multirow{6}{*}{Gender}} & -3                      & -0.040                     &-0.113  & -0.234                     &     -0.108                 &-0.188  \\
\multicolumn{1}{l|}{}                        & -2                      & -0.024                     &-0.064  & -0.165                     &    -0.098                  &-0.118  \\
\multicolumn{1}{l|}{}                        & -1                      & -0.012                     &-0.032  & -0.087                     &    -0.026                  & -0.046 \\
\multicolumn{1}{l|}{}                        & 1                       &  0.009                    &0.022  & 0.070                     &        0.058              & 0.038 \\
\multicolumn{1}{l|}{}                        & 2                       &  0.023                    &0.042  & 0.129                     &       0.090               &0.070  \\
\multicolumn{1}{l|}{}                        & 3                       &  0.029                    &0.068  & 0.178                     &       0.118               &0.104  \\ \hline
\end{tabular}}
\caption{Quantitative evaluation of editable. We measure the amount of age edit and gender edit when applying the same editing magnitude $\alpha$ on each method.}
\label{tab:edit}
\vspace{-10pt}
\end{table}

\noindent \textbf{Qualitative Results} \quad
We demonstrate the qualitative results for editing in Fig. \ref{fig:edit}. As our reconstruction is more robust than encoder-based methods, it allows better editing results. As can be seen, our work can perform reasonable edits while still preserving faithfully non-editing. $W^+$ space methods (pSp and StyleTransformer) invert the input image into poorly-behaved latent regions. Therefore, their editing is less meaningful and introduces significant artifacts. In contrast, our method produces significant editing effects with fewer artifact, which demonstrates the superior editing ability in the well-behaved $\mathcal{W}$ space. Specially, we also compare our method with previous works in the supplementary material, which utilize StyleGAN as generator. It is obvious that StyleGAN struggle to synthesize images from different viewing points, while our 3D-based GAN editing method can easily deal with different poses.

\subsection{Ablation Study}
\noindent \textbf{Is meta-auxiliary network required?} \quad
The row (2) in Tab. \ref{tab:ablation} demonstrates the importance of our meta-auxiliary network. To be specific, meta-auxiliary network aims to adapt the parameters of the generator to the given image, then experiment of row (2) directly fine-tune the parameters. As can be seen, the meta-auxiliary network can significantly accelerate optimization process and promote reconstruction performance.

\noindent \textbf{Is meta-learning required?} \quad
We test the effectiveness of the meta-learning strategy as row (3) in Tab. \ref{tab:ablation}. In this experiment, we directly train the auxiliary network without the subsequently meta-learning process. During inference, we fine-tune the parameters of the auxiliary network. The results listed in row (1) and (3) show that our full version with meta-learning surpasses the simplified ones.

\noindent \textbf{How important are the modules in Auxiliary network?} \quad
In the framework of our auxiliary network, we design two domain-specific module: 2D-related module and 3D-related module. We separately investigate the effectiveness of the two modules in Tab. \ref{tab:ablation}. In row (4), we remove 2D-related module which mainly provide missing texture details. The result shows that the missing details significantly degrade reconstruction performance. In addition, we remove 3D-related module which mainly rectify structural misalignment in row (5). As can be seen, the 3D information also promote reconstruction result but not as important as 2D-related module.

\begin{table}[]
\centering
\setlength{\tabcolsep}{1.5mm}{
\begin{tabular}{lccccc}
\hline
No. & Method        & MSE & LPIPS & FID & Times(s) \\ \hline
(1) & Ours(full)    & 0.0148    & 0.0718      & 14.57    & 4.11         \\
(2) & w/o Aux.      & 0.0335 & 0.1916      & 26.26    & 3.69         \\
(3) & w/o Meta      & 0.0169    & 0.1181      & 17.22    & 4.11         \\
(4) & w/o 2D module & 0.0236    & 0.1257     & 17.05    &  4.01        \\
(5) & w/o 3D module & 0.0157    & 0.0759      & 15.12    &  3.79        \\ \hline
\end{tabular}}
\caption{Ablation study.}
\label{tab:ablation}
\vspace{-10pt}
\end{table}

\subsection{Additional Study}
\noindent \textbf{Generator Adaptation} \quad
Although our method builds on 3D-GAN generator, the meta-auxiliary network is also effective in previous 2D-GAN generator, e.g., StyleGAN. In order to demonstrate the superior of the proposed method, we take StyleGAN as generator and compared with previous 2D-specific inversion methods. Specially, we only preserve 2D-related module in our auxiliary network, while adopting the same meta-learning strategy. The detailed experiments are listed in the supplementary material.



\begin{figure}[!t]
\centering

\begin{minipage}{1.0\linewidth}
\centering
\includegraphics[width=1.0\textwidth]{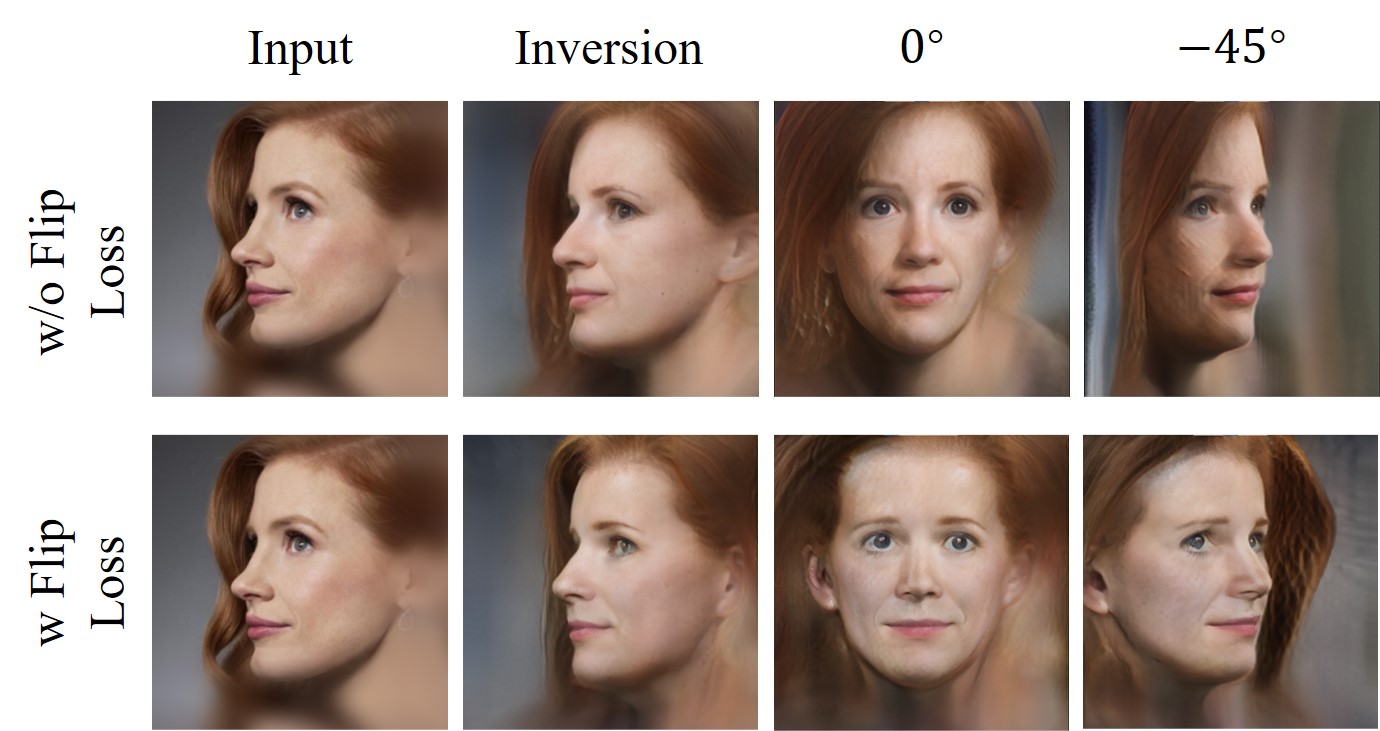}
\end{minipage}

\caption{Results of editing profile images. In the first row, we adopts Equ. \ref{equ:loss} as loss function. In the second row, we add an additional Flip Loss, as listed in Equ. \ref{equ:flip}.}
\label{fig:flip}
\end{figure}

\noindent \textbf{Editing Profile Images} \quad
Unlike 2D-GAN generator, 3D-GAN generator adopts 3D representation and contains information of 3D solid but not texture from a specific viewing point. However, GAN inversion normally takes a single image as input, which lacks comprehensive 3D information. The above issue would lead to unexpected results when editing profile images, as shown in the first row of Fig. \ref{fig:flip}. To deal with profile images, a simple but effective way is designing a loss function taking 3D structure into consideration. In this work, we have tried Flip Loss. Assume the input profile image is $I_P$, then we can obtain its horizontal flip image $I_P^{f}$. Obviously, the image $I_P^{f}$ is similar to the real image from the symmetrical viewing point. In addition, the 3D-GAN generator can easily obtain the inversion result from the corresponding symmetrical viewing point, denoted as $\hat{I}_P^{f}$. Then, we can use $I_P^{f}$ and $\hat{I}_P^{f}$ to construct a new loss function which take another viewing point into count. Considering $I_P^{f}$ may not be the same as the real image, the widely adopted $L_2$ loss is not suitable. In this work, we directly use LPIPS loss and GAN loss, which is the same as loss in Equ. \ref{equ:loss}. Then the overall loss function would be
\begin{equation}
\mathcal{L}_{f} = \lambda_{1}\mathcal{L}_{lpips}(I_P^{f}, \hat{I}_P^{f})+\lambda_{2}\mathcal{L}_{adv}(I_P^{f}, \hat{I}_P^{f})
\label{equ:flip}
\end{equation}
where $\lambda_1$ adn $\lambda_2$ is constants. The overall loss function is $\mathcal{L}_{all}=\mathcal{L}+\mathcal{L}_{f}$. With the Flip Loss, the reconstruction result is shown in the second row of Fig. \ref{fig:flip}. As we can seen, the Flip Loss introduce additional constraint on 3D structure and significantly promote editing performance of profile image.

\section{Conclusion}
In this paper, we introduce meta-auxiliary network, a novel framework for GAN inversion. We leverage meta-learning training strategy to accelerate optimization process, which achieve optimization-level reconstructions at encoder-like inference times. The proposed method successfully reduce performance and efficiency gap between encoder-based method and optimization-based method. In addition, we introduce 3D-GAN generator into GAN inversion, which can preserve identity consistency when rotate viewing point. Our method can edit facial attributes and rotate pose simultaneously, which remains a difficult issue for 2D-GAN inversion. Specially, the proposed method performs well both in 2D-GAN and 3D-GAN, demonstrating excellent generalization ability.
In summary, we believe this approach to be an essential step towards interactive and semantic in-the-wild image editing and may open the door for many intriguing real-world scenarios.

{\small
\bibliographystyle{ieee_fullname}
\bibliography{egbib}
}

\end{document}